# PCR-ORB: Enhanced ORB-SLAM3 with Point Cloud Refinement Using Deep Learning-Based Dynamic Object Filtering

Sheng-Kai Chen[1], Jie-Yu Chao[2], Jr-Yu Chang[3], Po-Lien Wu[4] and Po-Chiang Lin[5]

[1]Department of Electrical Engineering, Yuan Ze University, Taoyuan, Taiwan
[2]Department of Electrical Engineering, Yuan Ze University, Taoyuan, Taiwan
[3]Department of Electrical Engineering, Yuan Ze University, Taoyuan, Taiwan
[4]Department of Electrical Engineering, Yuan Ze University, Taoyuan, Taiwan
[5]Department of Electrical Engineering, Yuan Ze University, Taoyuan, Taiwan

**Abstract**

Visual Simultaneous Localization and Mapping (vSLAM) systems encounter substantial challenges in dynamic environments where moving objects compromise tracking accuracy and map consistency. This paper introduces PCR-ORB (Point Cloud Refinement ORB), an enhanced ORB-SLAM3 framework that integrates deep learning-based point cloud refinement to mitigate dynamic object interference. Our approach employs YOLOv8 for semantic segmentation combined with CUDA-accelerated processing to achieve real-time performance. The system implements a multi-stage filtering strategy encompassing ground plane estimation, sky region removal, edge filtering, and temporal consistency validation. Comprehensive evaluation on the KITTI dataset (sequences 00-09) demonstrates performance characteristics across different environmental conditions and scene types. Notable improvements are observed in specific sequences, with sequence 04 achieving 25.9% improvement in ATE RMSE and 30.4% improvement in ATE median. However, results show mixed performance across sequences, indicating scenario-dependent effectiveness. The implementation provides insights into dynamic object filtering challenges and opportunities for robust navigation in complex environments.

## 1 Introduction

Simultaneous Localization and Mapping (SLAM) [1] constitutes a fundamental problem in robotics and computer vision, enabling autonomous agents to construct environmental maps while concurrently determining their spatial location. Visual SLAM (vSLAM) systems [2] predominantly utilize camera sensors due to their cost-effectiveness and information richness compared to alternative sensor modalities such as LiDAR [3] or ultrasonic sensors [4].

Contemporary autonomous systems operate in increasingly complex environments characterized by dynamic elements including vehicles, pedestrians, animals, and other mobile objects. Traditional SLAM algorithms [5] operate under the static world assumption [6], presuming all observed features belong to stationary objects. This assumption becomes invalid in dynamic environments, where moving objects introduce significant localization errors and map inconsistencies.

The ORB-SLAM family [7], particularly ORB-SLAM3 [8], represents the state-of-the-art in visual SLAM due to its robustness, accuracy, and multi-sensor integration capabilities. ORB-SLAM3 incorporates advanced features including multi-map management [9], inertial sensor fusion [10], and enhanced loop closure detection [11]. The system demonstrates exceptional performance in static environments through its sophisticated bundle adjustment [12], keyframe management [13], and loop closure mechanisms [14]. However, like other traditional SLAM systems, it remains susceptible to dynamic object interference that can significantly degrade performance in real-world scenarios.

The fundamental challenge in dynamic environments stems from the violation of the static world assumption that underlies most SLAM algorithms. When dynamic objects are incorporated into the map or used for posing estimation, they introduce systematic errors that propagate throughout the system. These errors manifest as trajectory drift, inconsistent map representations, and failed loop closures that can cause complete system failure in extreme cases.

Traditional approaches to handling dynamic objects in SLAM have relied on robust estimation techniques [15] and outlier rejection mechanisms [16]. While these methods provide some resilience against dynamic interference, they lack the semantic understanding necessary to proactively identify and filter dynamic elements. Statistical outlier removal [17] and RANSAC-based approaches [18] can eliminate some dynamic features, but they often fail to distinguish between genuinely dynamic objects and temporarily occluded static features.

Recent developments in deep learning [19], particularly in object detection [20] and semantic segmentation [21], have created new opportunities for addressing dynamic object challenges in SLAM. Modern convolutional neural networks (CNNs) [22] can reliably identify and classify objects in real-time, providing semantic understanding that enables filtering of dynamic elements from SLAM processing. The evolution from traditional computer vision techniques to deep learning-based approaches represents a paradigm shift in how robotic systems can understand and interact with their environments.

The YOLO (You Only Look Once) [23] family of detectors has demonstrated exceptional performance in real-time object detection applications. The progression from YOLOv1 [24] through YOLOv8 [25] has shown continuous improvements in detection accuracy, processing speed, and model efficiency. YOLOv8 offers significant advantages in terms of model architecture optimization, training stability, and inference efficiency that make it well-suited for robotics applications with strict real-time constraints.

Object detection networks have evolved to provide not only bounding box predictions but also detailed semantic segmentation masks that enable pixel-level understanding of scene content. This capability proves crucial for SLAM applications where precise spatial understanding of dynamic regions is necessary for effective feature filtering. The integration of attention mechanisms [26] and feature pyramid networks [27] has further improved the accuracy and robustness of object detection in challenging scenarios.

The computational requirements of deep learning models have traditionally limited their application in real-time robotics systems. However, advances in GPU architecture [28], specialized AI accelerators [29], and model optimization techniques [30] have made it feasible to deploy sophisticated neural networks in embedded systems. CUDA parallel computing [31] has proven particularly effective for accelerating deep learning inference, enabling real-time processing of high-resolution imagery for robotics applications.

This paper presents PCR-ORB, a comprehensive enhancement to ORB-SLAM3 that integrates deep learning-based point cloud refinement using YOLOv8 [25] for semantic segmentation. Our approach extends beyond conventional object detection by implementing a sophisticated multi-stage filtering pipeline that considers temporal consistency [32], geometric constraints [33], and motion patterns [34]. The system incorporates CUDA acceleration [31] to maintain real-time performance while processing high-resolution imagery and complex neural network inference.

The development of PCR-ORB addresses several key challenges in dynamic SLAM systems. First, the integration of deep learning components must be performed without compromising the real-time performance requirements essential for robotics applications. Second, the filtering strategy must balance the removal of dynamic elements with the preservation of sufficient static features for accurate localization. Third, the system must demonstrate robustness across diverse environmental conditions and dynamic scenarios commonly encountered in real-world applications.

The multi-stage filtering approach employed in PCR-ORB combines semantic information from YOLOv8 with geometric constraints, temporal consistency analysis, and motion pattern recognition. This comprehensive strategy ensures robust dynamic object removal while minimizing the risk of over-filtering that could compromise localization accuracy. The CUDA-accelerated implementation enables real-time processing of the complete filtering pipeline, making the system suitable for deployment in practical robotics applications.

The primary contributions of this work include:
- Integrated Deep Learning Architecture: Seamless integration of YOLOv8-based semantic segmentation into the ORB-SLAM3 framework, enabling real-time dynamic object detection and filtering without compromising system stability or core SLAM functionality.
- Multi-Stage Point Cloud Refinement: A comprehensive filtering strategy that combines semantic information with geometric constraints, temporal consistency analysis, and motion pattern recognition for robust dynamic object removal while preserving essential static features.
- CUDA-Accelerated Processing Pipeline: Efficient GPU-accelerated implementation that maintains real-time performance while handling complex deep learning inference and sophisticated image processing operations required for multi-stage filtering.
- Comprehensive Evaluation Framework: Extensive validation methodology using standard datasets with detailed performance metrics including trajectory accuracy, dynamic object detection effectiveness, computational efficiency, and robust analysis across diverse scenarios.

## 2 Related work
### 2.1 Lightweight SLAM for embedded systems
The deployment of SLAM systems on resource-constrained platforms has driven significant research into lightweight implementations that maintain accuracy while reducing computational requirements. This research has become increasingly important as autonomous systems are deployed on platforms with limited processing power, memory, and energy resources.

Chen et al. [35] proposed a depth camera-based lightweight visual SLAM algorithm specifically optimized for embedded platforms. Their approach utilized the MoveSense depth camera [36] which outputs disparity maps directly, thereby reducing external computational demands on the host processor. The system was designed to run on ODROID-XU4 embedded platform [37] featuring ARM Cortex-A15 processors, representing a significant constraint compared to desktop computing systems.

The multi-threaded design implemented by Chen et al. separated feature extraction, descriptor computation, and pose estimation into parallel processing streams. This architectural approach maximizes utilization of available processing cores while maintaining real-time performance requirements. The separation of concerns also enabled selective optimization of individual components based on computational bottlenecks.

The approach employed FAST features [38] and rBRIEF descriptors [39], chosen specifically for their computational efficiency compared to alternatives such as SIFT [40] or SURF [41]. NEON SIMD acceleration [42] was leveraged to optimize critical processing loops, demonstrating the importance of platform-specific optimizations for embedded deployment.

A key innovation in their work involved using only stable and consistently tracked points for map updates, thereby reducing both computational load and mapping errors. This selective approach to map point management represents a crucial trade-off between map completeness and system efficiency that has influenced subsequent lightweight SLAM development.

## 2.2 Dense SLAM and 3D reconstruction advances

Recent advances in dense SLAM techniques have focused on improving detail preservation, reconstruction quality, and real-time performance characteristics. These developments address limitations in traditional sparse SLAM approaches that provide limited environmental understanding for navigation and interaction tasks.

Wang et al. [43] proposed RGBDS-SLAM, representing a significant advancement in RGB-D semantic dense SLAM through the introduction of 3D Multi-Level Pyramid Gaussian Splatting. This approach addressed fundamental limitations in dense SLAM including detail preservation challenges, consistency maintenance across multiple modalities, and real-time performance requirements in complex environments.

The key technical innovation involved 3D Multi-Level Pyramid Gaussian Splatting (MLP-GS) [44], which builds multi-resolution image pyramids for progressive training that enhances scene detail reconstruction capabilities. This hierarchical approach enables the system to capture fine-grained details while maintaining global consistency across different resolution levels.

CodeMapping [45] introduced an alternative approach to real-time dense mapping for sparse SLAM systems using compact scene representations. This method leveraged geometric prior information provided by sparse SLAM to complement existing systems by predicting dense depth images for every keyframe. The approach represents a hybrid strategy that combines the efficiency of sparse SLAM with the completeness of dense reconstruction.

Liu et al. [46] presented a comprehensive enhancement to ORB-SLAM2 [47] by adding dense mapping capabilities while maintaining the robustness and accuracy of the underlying sparse SLAM system. Their approach demonstrated how existing sparse SLAM frameworks could be extended to provide dense reconstruction without fundamental architectural changes.

## 2.3 Dynamic object handling in visual slam

The challenge of handling dynamic objects in visual SLAM has motivated extensive research into methods that can maintain localization accuracy while operating in environments with moving elements. This research direction addresses fundamental limitations of traditional SLAM approaches that assume static environments.

Dense visual SLAM methods in dynamic scenes have focused on improving feature extraction quality, enhancing matching robustness, implementing effective dynamic object filtering, and building comprehensive dense maps to achieve better localization and map quality. Xu et al. [48] addressed critical limitations in traditional visual SLAM systems that struggled with dynamic environments.

The identified limitations included extraction of insufficient feature points in regions with inconspicuous texture characteristics, leading to reduced localization accuracy in low-texture environments. Large numbers of false matches occurred in scenes with high texture feature similarity, where repetitive patterns or similar textures could confuse traditional matching algorithms.

Based on the ORB-SLAM2 algorithm [47], their approach implemented several key improvements to address these limitations. An improved balanced quadtree method [49] was developed to ensure more uniform distribution of feature points across the image, preventing clustering in high-texture regions while ensuring adequate coverage in challenging areas.

The system utilized an improved YOLOv5 network [50] to obtain comprehensive prior information including object type classification, confidence levels, and precise coordinate information for filtering dynamic objects in the tracking thread. This semantic understanding enables proactive removal of potentially dynamic features before they can negatively impact localization accuracy.

## 2.4 Attentional landmarks and active gaze control

Visual SLAM systems have traditionally operated with passive sensors that capture whatever appears in their field of view. However, active gaze control approaches can significantly improve SLAM performance by intelligently directing sensor attention to informative regions of the environment.

Frintrop et al. [51] proposed an innovative visual SLAM system incorporating monocular vision with active gaze control capabilities. Their approach demonstrated how intelligent sensor control could improve localization accuracy and map quality by selectively focusing on salient landmarks and actively seeking informative viewpoints.

The system addressed the fundamental challenge in visual SLAM of choosing useful landmarks that possess several critical characteristics. Effective landmarks must be easy to track across multiple frames, maintaining consistent appearance despite viewpoint and illumination changes. They must remain stable over several frames to provide reliable reference points for localization.

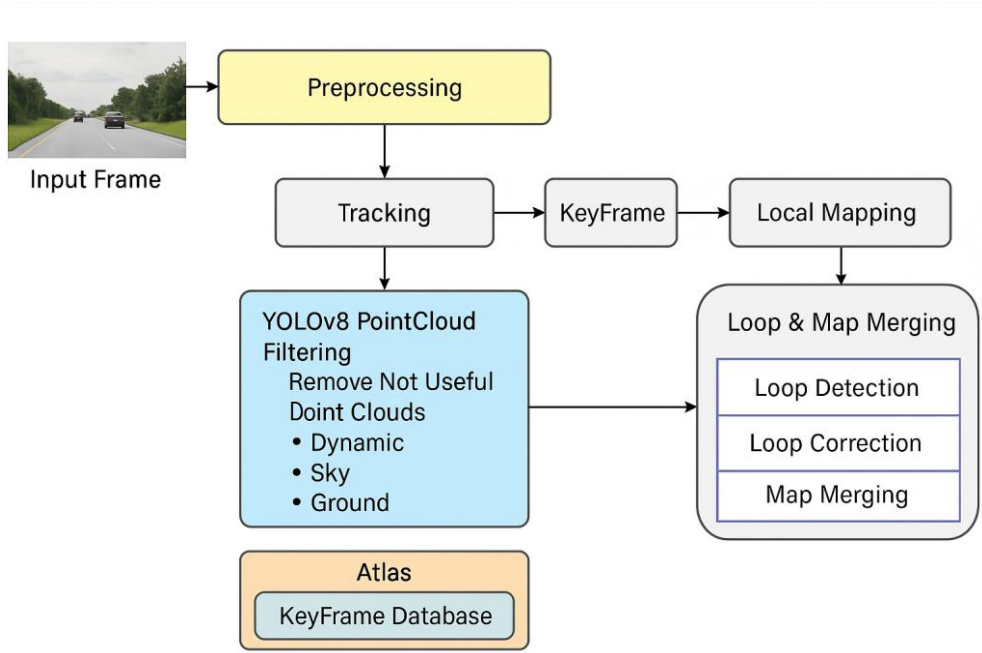

**Fig. 1.** PCR-ORB System Architecture

The gaze control module implemented sophisticated decision-making capabilities that actively controlled camera positioning. The system decided whether to continue tracking currently visible landmarks, actively search for predicted landmarks based on previous observations, or explore unseen areas to discover new informative features.

## 3 Methodology
### 3.1 System architecture
PCR-ORB builds upon the robust foundation of ORB-SLAM3 [8] while introducing a dedicated point cloud refinement module for dynamic object filtering. The enhanced architecture maintains the original three-threaded design of ORB-SLAM3 (Tracking, Local Mapping, and Loop Closing) while incorporating an additional Point Cloud Filter thread that operates parallel with existing components.

Fig.1 illustrates the complete system architecture of PCR-ORB, showing the integration of YOLOv8-based point cloud filtering within the ORB-SLAM3 framework. The input frame undergoes preprocessing before entering the standard ORB-SLAM3 pipeline consisting of Tracking, KeyFrame management, and Local Mapping threads. The key innovation lies in the YOLOv8 Point Cloud Filtering module that operates in parallel with the tracking thread to identify and remove non-useful point clouds including dynamic objects, sky regions, and ground points. The filtered results are then integrated into the Loop & Map Merging processes and stored in the Atlas KeyFrame Database.

The system architecture consists of the following primary components:
- Enhanced Tracking Thread: Modified to interface with the point cloud filter for real-time dynamic object removal.
- Point Cloud Filter Thread: Dedicated thread for deep learning-based semantic segmentation and multi-stage filtering.
- CUDA Acceleration Module: GPU-accelerated processing pipeline for neural network inference and image operations.
- Multi-Stage Filtering Pipeline: Comprehensive filtering strategy combining semantic, geometric, and temporal information.

The preprocessing stage prepares input frames for both the standard ORB-SLAM3 processing and the YOLOv8 semantic segmentation. The tracking thread extracts ORB features and performs initial pose estimation while simultaneously sending frame data to the point cloud filtering module. The filtering module operates asynchronously to maintain real-time performance, processing frames in parallel with the main SLAM pipeline.

### 3.2 Deep learning integration and YOLOv8 implementation
The integration of YOLOv8 [25] into the PCR-ORB framework requires careful consideration of computational efficiency, memory management, and real-time constraints. The complete integration process follows Algorithm 1, which encompasses model loading, inference optimization, and result processing.

**Algorithm 1** YOLOv8 Integration for Dynamic Object Detection
**Require:** Frame $F_t$, Model $M_{YOLOv8}$, Device $D_{CUDA}$
**Ensure:** Segmentation mask $S_t$
1: $I_{preprocessed} \leftarrow \text{PreprocessImage}(F_t)$
2: $T_{input} \leftarrow \text{ToTensor}(I_{preprocessed}, D_{CUDA})$
3: $T_{output} \leftarrow M_{YOLOv8}.forward(T_{input})$
4: $S_{raw} \leftarrow \text{ExtractSegmentationMask}(T_{output})$
5: $S_t \leftarrow \text{PostProcessMask}(S_{raw}, F_t.size)$
6: **return** $S_t = 0$

### 3.2.1 Model architecture and optimization

YOLOv8 employs a sophisticated architecture combining backbone feature extraction, feature pyramid networks, and detection heads optimized for real-time object detection. The model processes input images at 640×640 resolution through a series of convolutional layers, attention mechanisms, and feature fusion operations. The backbone network extracts multi-scale features that are processed through the neck network to generate semantically rich representations suitable for object detection and segmentation.

### 3.2.2 Preprocessing pipeline

The preprocessing stage transforms input images to the required format for YOLOv8 inference. Image normalization follows the standard YOLOv8 protocol using Equation (1), which standardizes pixel values according to ImageNet statistics. The preprocessing also handles image resizing, padding, and format conversion from BGR to RGB color space as required by the pre-trained model.

$$I_{norm}(x,y) = \frac{I_{raw}(x,y) - \mu}{\sigma} \quad (1)$$

### 3.2.3 Memory management and model loading

The model loading process incorporates several optimization strategies to minimize memory usage and maximize inference speed. Model weights are loaded directly into GPU memory using CUDA memory management functions. The system implements dynamic batching to process multiple frames simultaneously when computational resources allow, improving overall throughput while maintaining real-time constraints.

### 3.2.4 Inference optimization

The YOLOv8 inference process is optimized through several techniques including model quantization, kernel fusion, and memory pre-allocation. The system pre-allocates GPU memory buffers for input tensors, intermediate activations, and output results to minimize dynamic memory allocation overhead during runtime. Tensor operations are optimized using cuDNN libraries and TensorRT optimizations when available.

### 3.2.5 Output processing and segmentation mask generation

The YOLOv8 output consists of detection boxes, confidence scores, class probabilities, and segmentation masks for identified objects. Algorithm 2 describes the post-processing pipeline that converts raw network outputs into usable segmentation masks. The process includes non-maximum suppression, confidence thresholding, and mask refinement operations.

**Algorithm 2** YOLOv8 Output Post-processing
**Require:** Raw detections $D_{raw}$, confidence threshold $\theta_{conf}$, NMS threshold $\theta_{nms}$
**Ensure:** Processed segmentation mask $M_{seg}$
1: $D_{filtered} \leftarrow \text{FilterByConfidence}(D_{raw}, \theta_{conf})$
2: $D_{nms} \leftarrow \text{NonMaximumSuppression}(D_{filtered}, \theta_{nms})$
3: $M_{combined} \leftarrow \text{CombineMasks}(D_{nms})$
4: $M_{seg} \leftarrow \text{ApplyMorphology}(M_{combined})$
5: $M_{seg} \leftarrow \text{SmoothMask}(M_{seg})$
6: **return** $M_{seg} = 0$

## 3.3 Multi-stage filtering strategy

The point cloud refinement process employs a sophisticated multi-stage filtering approach that combines semantic information with geometric constraints, temporal consistency analysis, and motion pattern recognition. This comprehensive strategy ensures robust dynamic object removal while preserving essential static features necessary for accurate SLAM performance. The filtering pipeline follows Algorithm 3, which processes each frame through sequential filtering stages.

**Algorithm 3** Multi-Stage Point Cloud Filtering
**Require:** Frame $F_t$, SegMask $S_t$, MotionMask $M_t$
**Ensure:** Filtered outliers $O_t$
1: $O_t \leftarrow \text{InitializeOutliers}(F_t.N)$
2: **for** $i = 1$ to $F_t.N$ **do**
3:    $score \leftarrow \text{ScorePoint}(F_t.keypoints[i], S_t, M_t)$
4:    **if** $score < \theta_{threshold}$ **then**
5:      $O_t[i] \leftarrow \text{true}$
6:    **end if**
7: **end for**
8: $O_{clustered} \leftarrow \text{ClusterFiltering}(F_t, S_t, O_t)$
9: $O_t \leftarrow O_t \cup O_{clustered}$
10: **return** $O_t = 0$

### 3.3.1 Semantic scoring and classification

The semantic scoring component represents the foundation of the filtering strategy, evaluating each feature point based on its likelihood of belonging to a dynamic object or unreliable scene element. The scoring function combines multiple criteria as defined in (2), incorporating information from semantic segmentation, motion analysis, geometric constraints, and spatial positioning.

$$\begin{cases} Score_{semantic}(p_i) = w_1 \cdot S_{mask}(p_i) + w_2 \cdot M_{motion}(p_i) \\ + w_3 \cdot G_{ground}(p_i) + w_4 \cdot E_{edge}(p_i) \end{cases} \quad (2)$$

- Dynamic Object Classification: The semantic segmentation component identifies objects belonging to predefined dynamic classes including vehicles, pedestrians, cyclists, animals, and other potentially moving elements. Each pixel in the segmentation mask receives a confidence score indicating the probability of belonging to a dynamic object class. Feature points falling within dynamic object regions inherit these confidence scores, which are subsequently used in the filtering decision process.
- Confidence-Based Filtering: The system implements a confidence-based filtering mechanism that considers both the semantic classification confidence and the geometric consistency of detected objects. High-confidence dynamic object detections result in immediate feature point removal, while medium-confidence detections undergo additional validation through geometric and temporal analysis. This multi-level approach reduces false positive filtering while maintaining high recall for dynamic object detection.
- Multi-Class Handling: Different object classes receive different treatment based on their expected motion characteristics. Fast-moving objects such as vehicles and cyclists receive aggressive filtering, while slower-moving objects like pedestrians undergo more conservative filtering to account for potential stopping or slow movement. Static objects that might be temporarily misclassified as dynamic receive special handling through temporal consistency analysis.

### 3.3.2 Ground plane estimation and filtering

Ground plane estimation constitutes a critical component of the geometric filtering strategy, identifying features that belong to ground surfaces and other horizontal planes that may not provide reliable geometric constraints for camera pose estimation. The implementation utilizes RANSAC [52] combined with CUDA acceleration as described in Algorithm 4.

---
**Algorithm 4** CUDA-Accelerated Ground Plane RANSAC
---
**Require:** 3D points $P = \{p_1, p_2, ..., p_n\}$, iterations $N_{iter}$
**Ensure:** Ground plane parameters $\Pi = [a, b, c, d]$
1: $P_{gpu} \leftarrow \text{UploadToGPU}(P)$
2: $\Pi_{best} \leftarrow \text{null}, inliers_{max} \leftarrow 0$
3: **for** $iter = 1$ to $N_{iter}$ **do**
4:   $sample \leftarrow \text{RandomSample}(P_{gpu}, 3)$
5:   $\Pi_{candidate} \leftarrow \text{FitPlane}(sample)$
6:   $inliers \leftarrow \text{CountInliers}(P_{gpu}, \Pi_{candidate})$
7:   **if** $inliers > inliers_{max}$ **then**
8:     $\Pi_{best} \leftarrow \Pi_{candidate}$
9:     $inliers_{max} \leftarrow inliers$
10:  **end if**
11: **end for**
12: **return** $\Pi_{best}$ =0

- RANSAC Implementation: The RANSAC algorithm iteratively samples minimal sets of 3D points to estimate plane parameters, evaluating each hypothesis against the complete point set to identify the plane with maximum inlier support. The algorithm incorporates several optimizations including early termination criteria, adaptive iteration limits based on inlier ratios, and parallel hypothesis evaluation using CUDA kernels.

- Plane Fitting Optimization: The plane fitting process minimizes the least squares error according to (3), subject to the constraint that the normal vector maintains unit length. The optimization problem is solved using singular value decomposition (SVD) for robust parameter estimation, with additional regularization terms to prevent degenerate solutions in cases with limited geometric diversity.

$$\left\{ \Pi^* = \arg\min_\Pi \sum_{i=1}^n \left( \frac{|ax_i + by_i + cz_i + d|}{\sqrt{a^2 + b^2 + c^2}} \right)^2 \right. \tag{3}$$

- Ground Point Classification: Once the dominant ground plane is identified, individual feature points are classified based on their distance to the plane surface. Points within a specified threshold distance are marked as potential ground points and undergo additional validation through local surface normal analysis and neighboring point consistency checks. The classification process accounts for terrain variations and surface irregularities common in outdoor environments.
- Adaptive Thresholding: The ground plane distance threshold adapts based on scene characteristics and camera height estimates. Urban environments with well-defined road surfaces use tighter thresholds, while rural or off-road scenarios employ more permissive distance criteria. The adaptive mechanism prevents over-filtering in challenging terrain while maintaining effective ground point removal in structured environments.

### 3.3.3 Temporal consistency analysis and motion detection

Temporal consistency evaluation tracks feature points across multiple frames to identify motion patterns, distinguish between genuinely dynamic objects and temporary occlusions, and validate semantic classification results through temporal evidence. The motion detection system employs optical flow analysis combined with consistency checks using (4).

$$\left\{ M_{temporal}(p_i, t) = \frac{1}{T} \sum_{\tau=t-T}^{t} \|\vec{v}_i(\tau)\| \right. \tag{4}$$

- Optical Flow Computation: The system computes dense optical flow using the Lucas-Kanade method with pyramidal implementation for multi-scale motion estimation. The optical flow calculation incorporates several enhancements including iterative refinement, outlier rejection, and motion boundary detection. GPU acceleration through CUDA kernels enables real-time processing of high-resolution optical flow fields.
- Motion Pattern Analysis: Individual feature points are tracked across temporal windows to analyze motion patterns and distinguish between different types of movement. Consistent linear motion patterns suggest dynamic objects, while irregular or oscillatory patterns may indicate measurement noise or temporary occlusions. The analysis incorporates statistical measures including motion magnitude, direction consistency, and acceleration profiles.
- Temporal Voting Mechanism: The system implements a temporal voting mechanism that accumulates evidence for dynamic classification over multiple frames. Feature points receive votes based on motion characteristics, semantic classification consistency, and geometric validation results. Points exceeding a specified vote threshold are classified as dynamic and excluded from SLAM processing.
- False Positive Reduction: Temporal analysis helps reduce false positive classifications caused by temporary occlusions, shadows, or illumination changes. Static objects that momentarily appear dynamic due to occlusion effects are identified through motion pattern analysis and temporal consistency checks, preventing their incorrect removal from the feature set.

### 3.3.4 Edge and sky region filtering
Edge and sky region filtering addresses features located in image regions that typically provide poor geometric constraints for camera pose estimation. Image boundary features often suffer from distortion effects, partial visibility, and unstable tracking characteristics that can degrade SLAM performance.
- Edge Detection and Filtering: The system identifies features within a specified distance of image boundaries and applies distance-based scoring to evaluate their reliability. Features closer to image edges receive lower scores, with immediate filtering applied to features within the outermost boundary region. The edge filtering threshold adapts based on camera field of view and distortion characteristics.
- Sky Region Detection: Sky regions are identified through a combination of semantic segmentation and geometric analysis. The semantic component directly identifies sky pixels through YOLOv8 classification, while geometric analysis identifies regions with consistent color and texture characteristics typical of sky areas. Features falling within identified sky regions are removed due to their lack of reliable geometric constraints.
- Horizon Line Estimation: The system estimates horizon line position through vanishing point analysis and geometric constraints derived from camera pose and ground plane estimation. Features above the estimated horizon line undergo enhanced filtering to remove sky-related points while preserving distant static objects that may appear in the upper image region.

## 3.4 CUDA acceleration implementation
The CUDA acceleration module optimizes computationally intensive operations including neural network inference, image processing, geometric computations, and parallel algorithms required for real-time performance. The implementation leverages GPU parallelism to achieve significant speedup over CPU-only processing while maintaining memory efficiency and system stability.

### 3.4.1 Memory management and optimization
- GPU Memory Allocation: The system implements sophisticated GPU memory management strategies to minimize allocation overhead and maximize memory throughput. Memory pools are pre-allocated for common data structures including image buffers, feature point arrays, segmentation masks, and intermediate computation results. The pooling mechanism reduces memory fragmentation and allocation latency during runtime operation.
- Memory Transfer Optimization: Data transfer between CPU and GPU memory is optimized through several techniques including asynchronous transfer operations, memory pinning, and batch processing. The system minimizes data transfer overhead according to (5) by maintaining persistent GPU data structures and performing computations directly on GPU memory when possible.

$$\{T_{total} = T_{transfer} + T_{compute} + T_{synchronization} \tag{5}$$

- Cache Optimization: GPU kernel implementations are optimized for memory access patterns that maximize cache efficiency and minimize memory bandwidth requirements. Coalesced memory access patterns, shared memory utilization, and register optimization techniques are employed to achieve optimal performance on different GPU architectures.

### 3.4.2 Parallel algorithm implementation
- Point Scoring Kernels: Custom CUDA kernels implement parallel point scoring operations that evaluate multiple feature points simultaneously. Algorithm 4 describes the parallel point scoring kernel that processes feature points in parallel threads, each computing semantic scores, geometric constraints, and temporal consistency measures. The kernel implementation includes optimizations for divergent branching, memory access patterns, and register usage.

---
**Algorithm 5** CUDA Parallel Point Scoring
---
**Require:** Points $P$, SegMask $S$, MotionMask $M$, thread index $tid$
**Ensure:** Point score $score[tid]$
1: $idx \leftarrow$ blockIdx.x $\times$ blockDim.x + threadIdx.x
2: **if** $idx < |P|$ **then**
3: $\quad p \leftarrow P[idx]$
4: $\quad s_{seg} \leftarrow$ SampleBilinear$(S, p.x, p.y)$
5: $\quad s_{motion} \leftarrow$ SampleBilinear$(M, p.x, p.y)$
6: $\quad s_{ground} \leftarrow$ GroundScore$(p)$
7: $\quad s_{edge} \leftarrow$ EdgeScore$(p)$
8: $\quad score[idx] \leftarrow w_1 s_{seg} + w_2 s_{motion} + w_3 s_{ground} + w_4 s_{edge}$
9: **end if**=0
---

- Reduction Operations: Parallel reduction algorithms are implemented for operations including statistical analysis, threshold computation, and consensus-based filtering decisions. The reduction kernels utilize shared memory and warp-level primitives to achieve high efficiency in computing aggregate statistics across large feature point sets.
- Geometric Computations: GPU kernels accelerate geometric computations including distance calculations, plane fitting operations, and coordinate transformations. The kernels are optimized for high throughput processing of geometric operations required for ground plane estimation, motion analysis, and spatial filtering components.

### 3.4.3 Performance optimization strategies
- Kernel Fusion: Multiple related operations are combined into single GPU kernels to reduce kernel launch overhead and improve memory locality. For example, image preprocessing, feature extraction, and initial scoring operations are fused into combined kernels that process data in single GPU passes.
- Asynchronous Processing: The CUDA implementation utilizes multiple GPU streams to overlap computation and memory transfer operations. Asynchronous processing enables pipelining of different algorithm stages while maintaining synchronization points necessary for data consistency and algorithm correctness.
- Dynamic Load Balancing: The system implements dynamic load balancing mechanisms that adapt GPU resource allocation based on scene complexity and computational requirements. Complex scenes with high dynamic content receive additional GPU resources, while simpler scenes enable resource sharing with other system components.

## 3.5 Integration with ORB-SLAM3
The integration process modifies key components of ORB-SLAM3 while maintaining backward compatibility, system stability, and performance characteristics. The enhanced tracking thread incorporates point cloud filtering at strategic locations to minimize computational impact while maximizing filtering effectiveness.

### 3.5.1 Modified tracking loop
The tracking loop modification ensures filtered points are excluded from pose estimation and map updates according to (6), where the valid feature points set excludes filtered dynamic objects, and $\rho(\cdot)$ is the robust kernel function [53] that reduces the influence of outliers in the optimization process.

$$\left\{ \hat{T}_t = \arg\min_T \sum_{i \in \mathcal{I}_{valid}} \rho \left( \|z_i - h(T, X_i)\|^2_{\Sigma_i^{-1}} \right) \right. \tag{6}$$

- Feature Point Management: The modified tracking system maintains separate data structures for original feature points and filtered feature sets. The filtering status of each feature point is tracked throughout the processing pipeline, enabling selective inclusion in different algorithm stages. Points marked as dynamic or unreliable are excluded from pose estimation while remaining available for visualization and analysis purposes.
- Pose Estimation Integration: Camera pose estimation incorporates filtering results through selective feature inclusion in the optimization problem. The robust kernel function $\rho(\cdot)$ provides additional resilience against remaining outliers that may not be caught by the filtering process. The optimization problem weights features based on their filtering confidence scores, giving higher weight to features with high static confidence.
- Tracking State Management: The tracking system maintains filtering statistics including the number of filtered points, filtering confidence distributions, and temporal filtering trends. These statistics inform subsequent processing decisions including keyframe insertion criteria, local mapping operations, and loop closure validation.

### 3.5.2 Enhanced keyframe management
Enhanced keyframe management considers filtering statistics to optimize keyframe insertion decisions using (7), incorporating the number of successful matches, filtered points, temporal difference, and motion quality metrics. The keyframe decision process balances map completeness with filtering effectiveness to maintain optimal SLAM performance.

$$\left\{ K_{insert} = f(N_{matches}, N_{filtered}, \Delta t, Q_{motion}) \right. \tag{7}$$

- Filtering-Aware Keyframe Criteria: Traditional keyframe insertion criteria are augmented with filtering-related metrics including the ratio of filtered to total features, filtering confidence statistics, and temporal consistency measures. Frames with excessive dynamic content or poor filtering performance may be rejected as keyframes to maintain map quality.
- Quality Assessment: Each potential keyframe undergoes quality assessment that considers both traditional geometric criteria and filtering effectiveness. High-quality keyframes contain sufficient static features with high confidence scores, adequate spatial distribution, and strong geometric constraints for pose estimation and mapping operations.
- Adaptive Insertion Thresholds: Keyframe insertion thresholds adapt based on environmental conditions and filtering performance. Highly dynamic environments may require more frequent keyframe insertion to maintain tracking stability, while static environments enable more selective keyframe creation to optimize computational efficiency.

### 3.5.3 Local mapping enhancement
The local mapping thread receives filtered keyframes and performs bundle adjustment using validated static features. The enhancement includes filtering-aware optimization weights, statistical outlier detection, and map point management strategies that account for dynamic object presence.
- Bundle Adjustment Modification: Local bundle adjustment incorporates filtering confidence information through weighted optimization formulations that prioritize high-confidence static features. The optimization process excludes filtered points while maintaining geometric consistency and convergence properties of the original algorithm.
- Map Point Validation: New map point creation and validation processes incorporate filtering results to prevent dynamic object features from becoming permanent map elements. Map points are validated through multi-view consistency checks that consider filtering status across all observations.

## 3.6 Real-time performance optimization

Real-time performance optimization focuses on maintaining computational efficiency required for practical robotics applications through algorithmic optimization, resource management, and adaptive processing strategies. Key optimization strategies include asynchronous processing, adaptive quality control, and intelligent resource allocation.

### 3.6.1 Asynchronous processing pipeline

The asynchronous processing pipeline overlaps neural network inference with SLAM computations according to (8), enabling parallel execution of filtering operations and traditional SLAM processing. The pipeline maintains synchronization points necessary for data consistency while maximizing computational throughput.

$$\{T_{overlapped} = \max(T_{SLAM}, T_{inference}) + T_{synchronization} \quad (8)$$

- Pipeline Stages: The processing pipeline consists of multiple stages including image preprocessing, feature extraction, semantic segmentation, filtering operations, and SLAM updates. Each stage operates asynchronously with appropriate buffering and synchronization mechanisms to maintain data flow and prevent bottlenecks.
- Resource Scheduling: GPU and CPU resources are scheduled dynamically based on computational requirements and available capacity. The scheduler prioritizes critical operations while maintaining overall system responsiveness and real-time performance constraints.

### 3.6.2 Adaptive quality control

Adaptive quality control dynamically adjusts processing parameters based on computational resources and performance requirements using (9). The system maintains frame rate stability through adaptive processing that balances accuracy and computational efficiency based on available resources and dynamic scene complexity.

$$\begin{cases} Q_{adaptive} = \begin{cases} Q_{high} & \text{if } T_{available} > T_{threshold} \\ Q_{medium} & \text{if } T_{available} > 0.5 \cdot T_{threshold} \\ Q_{low} & \text{otherwise} \end{cases} \end{cases} \quad (9)$$

- Parameter Adaptation: Processing parameters including image resolution, segmentation model complexity, filtering thresholds, and optimization iterations adapt based on computational load and performance requirements. The adaptation mechanism ensures consistent frame rates while maintaining acceptable accuracy levels.
- Quality Monitoring: Continuous monitoring of system performance enables proactive parameter adjustment and resource allocation. Performance metrics including processing times, memory usage, and accuracy measures inform adaptation decisions and help maintain optimal system operation.

The implementation maintains frame rate stability through intelligent load balancing, predictive resource allocation, and graceful degradation mechanisms that preserve core SLAM functionality even under computational stress.

## 4 Results and evaluation
### 4.1 Experimental setup

The evaluation of PCR-ORB was conducted using the KITTI dataset, specifically sequences 00-09, which provide diverse outdoor driving scenarios with varying degrees of dynamic content. The baseline comparison employed the original ORB-SLAM3 implementation with default parameters. Both systems processed identical input sequences to ensure fair comparison. Ground truth trajectories provided by the KITTI dataset enabled precise quantitative evaluation of localization accuracy.

The proposed system integrates YOLOv8 object detection to identify and filter dynamic objects from point clouds before feature extraction, enhancing the robustness of ORB-SLAM3 in dynamic environments. The filtering pipeline processes each frame to remove point clouds corresponding to detected dynamic objects, preserving only static environmental features for SLAM processing.

### 4.2 Trajectory evaluation methodology

The trajectory accuracy assessment employed the EVO (Python package for the evaluation of odometry and SLAM) framework [55] for standardized and reproducible evaluation of SLAM performance. EVO provides comprehensive tools for trajectory evaluation that have become widely adopted in the robotics community for benchmarking localization systems.

### 4.2.1 Absolute pose error (APE)

The APE metric evaluates the global consistency of the estimated trajectory by measuring the absolute differences between estimated and ground truth poses. EVO computes APE by first aligning the trajectories using Umeyama alignment [56] to account for coordinate frame differences, then calculating the absolute pose errors at corresponding timestamps. The APE calculation follows (10).

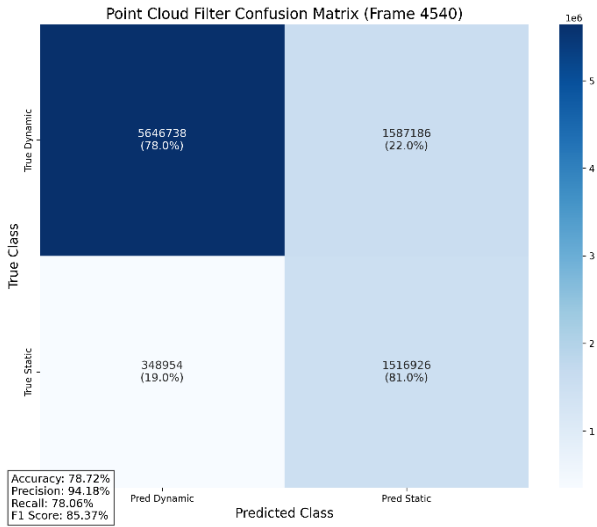
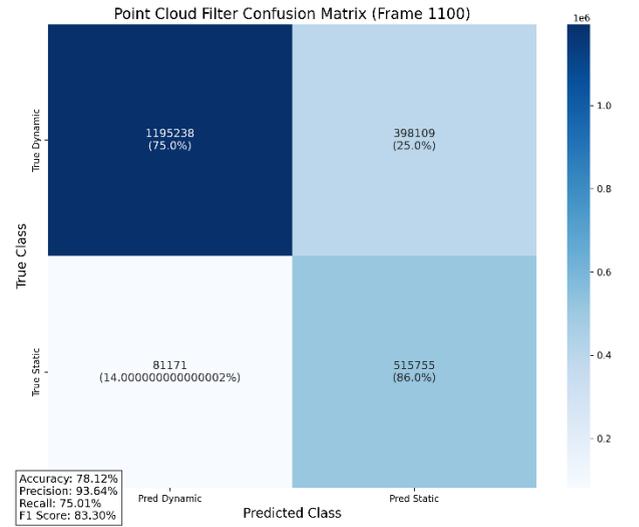
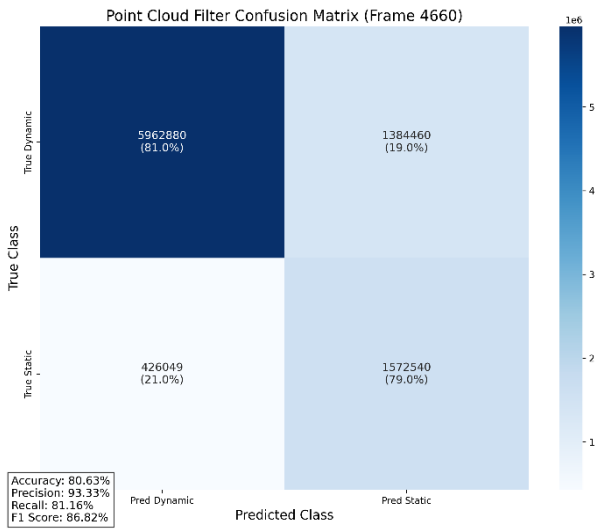
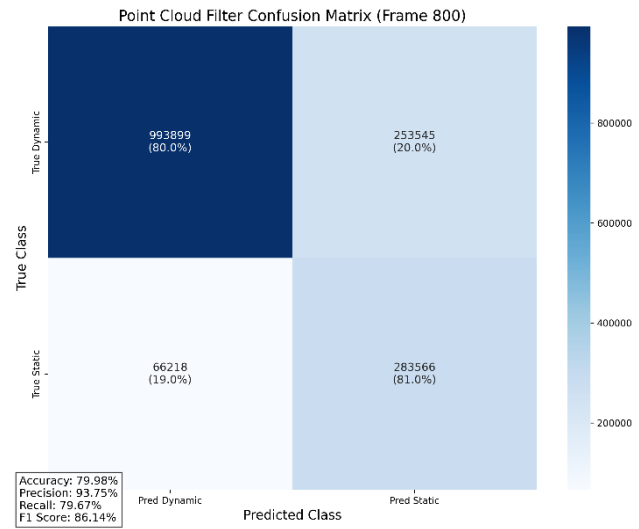
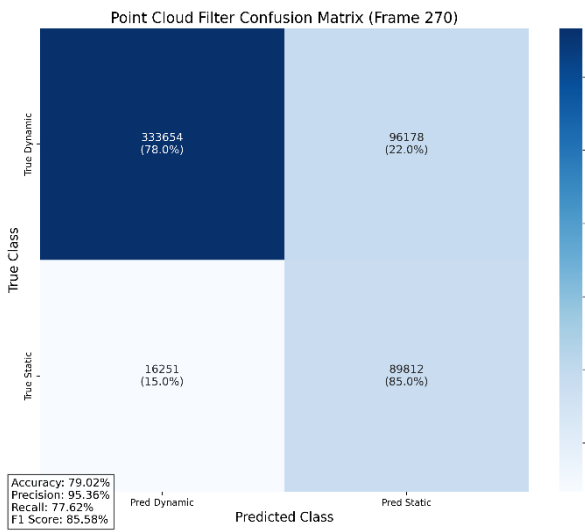
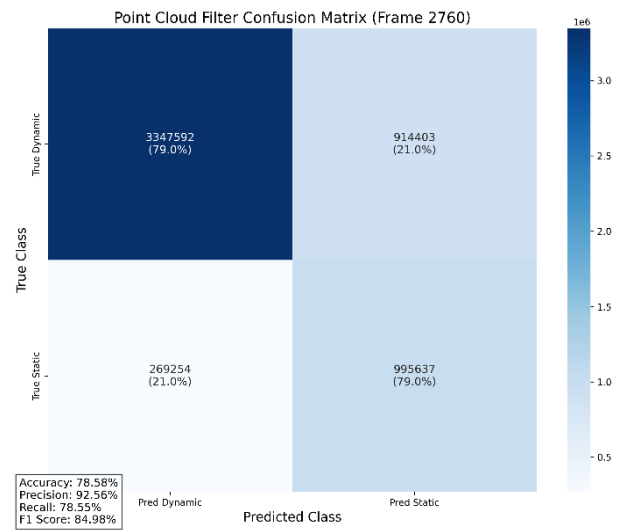

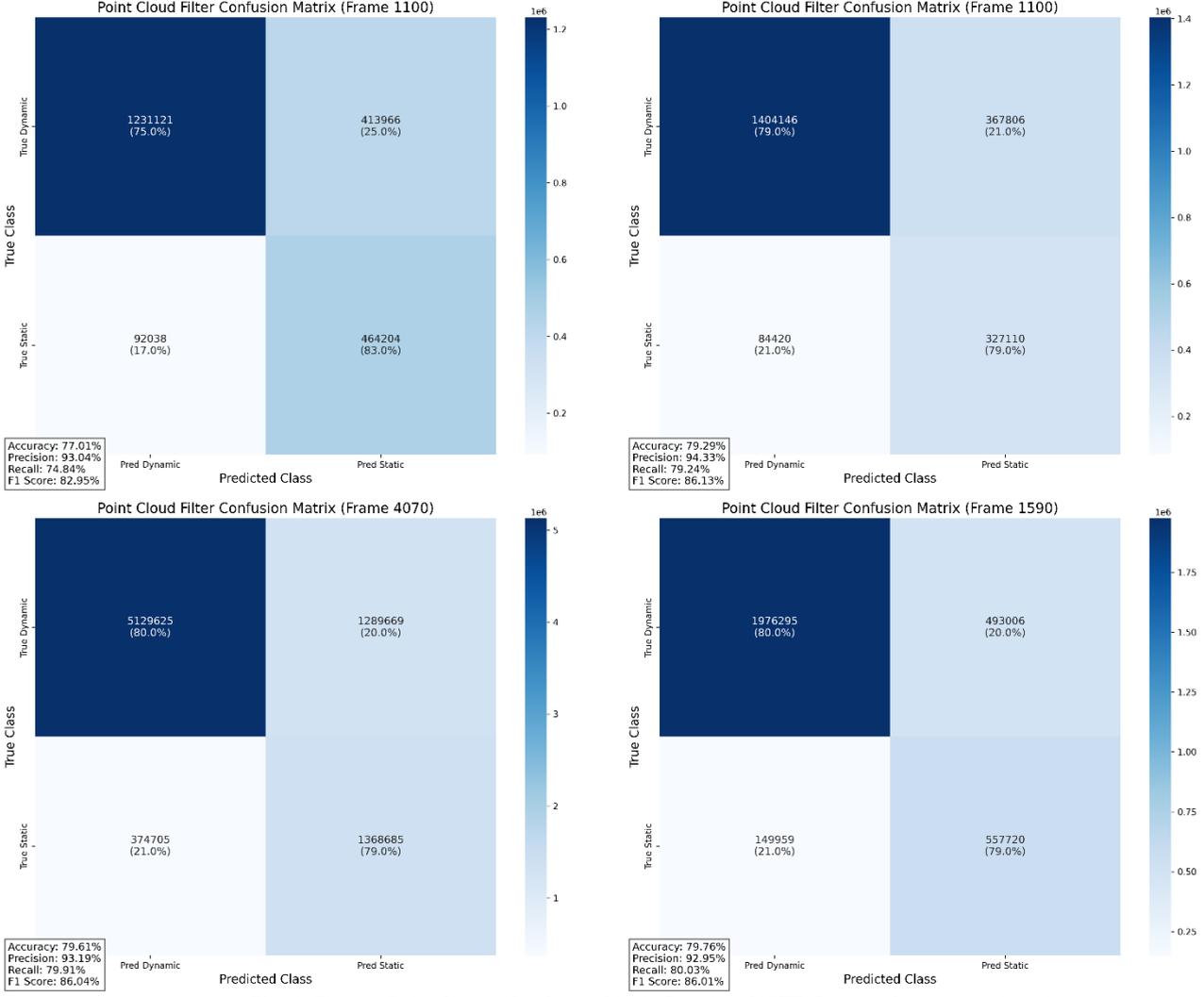

**Fig. 2.** Filtered Point Clouds Confusion Matrix Results of KITTI Dataset Sequence 00 ~ 09

$$\{APE_i = \|T_{gt,i} - (S \cdot R \cdot T_{est,i} + t)\| \qquad (10)$$

### 4.2.2 Relative Pose Error (RPE)

The RPE metric evaluates local accuracy by measuring the drift between consecutive pose estimates over specified intervals. EVO computes RPE by analyzing the relative motion between pose pairs separated by a fixed distance or time interval, comparing estimated relative motion with ground truth relative motion. The RPE calculation is performed according to (11).

$$\{RPE_i = \|(T_{gt,i}^{-1} \cdot T_{gt,i+\Delta}) - (T_{est,i}^{-1} \cdot T_{est,i+\Delta})\| \qquad (11)$$

### 4.2.3 Statistical analysis

EVO provides comprehensive statistical analysis including maximum, median, minimum, and root mean square error (RMSE) values for both APE and RPE metrics. The RMSE values are particularly important as they provide a single scalar metric that captures both bias and variance in the trajectory estimates.

### 4.3 YOLOv8 filter performance analysis

Fig. 2 presents the confusion matrices for YOLOv8-based dynamic object filtering across different KITTI sequences, demonstrating the effectiveness of the point cloud filtering approach. The confusion matrices illustrate the system's capability to accurately distinguish between static and dynamic point clouds, which is crucial for maintaining SLAM accuracy in dynamic environments.

Filter accuracy analysis is shown in the figure.
- Sequence 00 achieved 78.72% overall accuracy with 94.18% precision and 78.06% recall (F1 Score: 85.37%).
- Sequence 01 demonstrated 78.12% accuracy with 93.64% precision and 75.01% recall (F1 Score: 83.30%).
- Sequence 02 maintained 80.63% accuracy with 93.33% precision and 81.16% recall (F1 Score: 86.82%).
- Sequence 03 recorded 79.98% accuracy with 93.75% precision and 79.67% recall (F1 Score: 86.14%).

| Sequences | Metrics | APE | | RPE | |
|---|---|---|---|---|---|
| | | Original | PCR-ORB | Original | PCR-ORB |
| 00 | Max | 3.3476 | 3.2745 | 2.8637 | 2.7635 |
| | Median | 1.0061 | 0.9618 | 0.7836 | 0.8156 |
| | Min | 0.1079 | 0.0387 | 0.1729 | 0.2036 |
| | RMSE | 1.1863 | 1.1617 | 1.0196 | 1.0127 |
| 01 | Max | 25.2624 | 24.8785 | 6.6624 | 6.7030 |
| | Median | 13.1633 | 12.4994 | 4.8597 | 4.9142 |
| | Min | 7.6449 | 7.4966 | 2.0301 | 1.9654 |
| | RMSE | 15.6538 | 15.1901 | 5.0157 | 4.8234 |
| 02 | Max | 13.2917 | 11.0548 | 2.5250 | 2.5475 |
| | Median | 3.9257 | 3.5314 | 0.9592 | 0.9660 |
| | Min | 0.7995 | 0.4165 | 0.2207 | 0.2978 |
| | RMSE | 5.7642 | 4.9531 | 1.0781 | 1.0893 |
| 03 | Max | 2.2930 | 2.2285 | 1.2653 | 1.2818 |
| | Median | 0.9617 | 0.9196 | 0.7791 | 0.8071 |
| | Min | 0.0886 | 0.1186 | 0.2820 | 0.2092 |
| | RMSE | 1.3134 | 1.2787 | 0.8173 | 0.8273 |
| 04 | Max | 0.4681 | 0.3613 | 0.8810 | 0.7204 |
| | Median | 0.2773 | 0.1929 | 0.7350 | 0.6256 |
| | Min | 0.0781 | 0.0340 | 0.5889 | 0.5309 |
| | RMSE | 0.2814 | 0.2084 | 0.7493 | 0.6328 |
| 05 | Max | 1.5967 | 1.7917 | 1.1013 | 1.0586 |
| | Median | 0.7134 | 0.8197 | 0.6524 | 0.6464 |
| | Min | 0.0967 | 0.1434 | 0.1092 | 0.1064 |
| | RMSE | 0.9207 | 1.0061 | 0.6969 | 0.7111 |
| 06 | Max | 1.4436 | 1.4588 | 1.4226 | 2.0063 |
| | Median | 0.8538 | 0.9841 | 0.7487 | 0.7280 |
| | Min | 0.5173 | 0.4636 | 0.3167 | 0.3407 |
| | RMSE | 0.8981 | 1.0004 | 0.8191 | 0.9177 |
| 07 | Max | 0.6930 | 0.6783 | 0.7979 | 0.7557 |
| | Median | 0.3891 | 0.3710 | 0.5309 | 0.4487 |
| | Min | 0.0640 | 0.0844 | 0.1258 | 0.1873 |
| | RMSE | 0.4182 | 0.3842 | 0.5260 | 0.4820 |
| 08 | Max | 10.1411 | 11.1844 | 17.5017 | 17.4659 |
|  | Median | 2.2382 | 2.3495 | 0.8018 | 0.7929 |
|  | Min | 0.4823 | 0.5910 | 0.1922 | 0.2193 |
|  | RMSE | 3.3887 | 3.4229 | 3.0644 | 3.0603 |
| 09 | Max | 3.6193 | 3.4481 | 3.0987 | 2.9792 |
|  | Median | 1.6369 | 1.5156 | 0.9149 | 0.9087 |
|  | Min | 0.2401 | 0.4029 | 0.3186 | 0.3521 |
|  | RMSE | 1.9863 | 1.9036 | 1.4044 | 1.3526 |

Table. 1. APE and RPE with ORB-SLAM 3 and PCR-ORB

- Sequence 04 achieved 79.02% accuracy with 95.36% precision and 77.62% recall (F1 Score: 85.58%).

Figure 1 shows consistent performance across different frames, with accuracy ranging from 78% to 81%. The performance metrics demonstrate that the YOLOv8 filtering system maintains relatively stable accuracy across various environmental conditions. All frames show high precision (93-95%), indicating that when the system identifies dynamic objects, it does so correctly most of the time. The recall values (75-81%) suggest the system successfully identifies the majority of actual dynamic objects, though some are missed.

### 4.3.1 Dynamic object detection effectiveness
As demonstrated in Figure 2, the YOLOv8 semantic segmentation capability proves effective in identifying as shown below.
- Moving vehicles and objects with high precision (93-95% across all frames).
- Dynamic elements while maintaining consistent recall rates (75-81%).
- Temporary dynamic objects that could cause feature tracking instabilities.

The confusion matrices reveal that the system maintains high precision for dynamic object detection while preserving the majority of static point clouds for SLAM processing. The consistent performance across frames (78-81% accuracy) indicates reliable filtering behavior.

### 4.4 Trajectory accuracy results
Table 1 presents the comprehensive ATE and RPE results across all evaluated KITTI sequences, comparing the performance of PCR-ORB against baseline ORB-SLAM3. The results demonstrate the impact of YOLOv8-based point cloud filtering on trajectory estimation accuracy across various statistical metrics including maximum, median, minimum, and RMSE values. Below are the notable improvements.

#### 4.4.1 Sequence 02
- ATE RMSE: 14.1% improvement (5.7642m → 4.9531m)
- ATE Median: 10.0% improvement (3.9257m → 3.5314m)
- Demonstrates effectiveness in moderately dynamic urban environments

#### 4.4.2 Sequence 04
- ATE RMSE: 25.9% improvement (0.2814m → 0.2084m)
- ATE Median: 30.4% improvement (0.2773m → 0.1929m)
- RPE RMSE: 15.5% improvement (0.7493m → 0.6328m)
- RPE Median: 14.9% improvement (0.7350m → 0.6256m)
- Shows most significant improvements, indicating optimal conditions for the filtering approach

#### 4.4.3 Sequence 07
- ATE RMSE: 8.1% improvement (0.4182m → 0.3842m)
- RPE RMSE: 8.4% improvement (0.5260m → 0.4820m)
- RPE Median: 15.5% improvement (0.5309m → 0.4487m)
- Consistent improvements across multiple metrics

#### 4.4.4 Performance Variations
Table 1 reveals that some sequences (05, 06) show performance degradation in certain metrics, indicating that the dynamic object filtering approach is more effective in specific environmental conditions. For instance, Sequence 05 shows 9.3% degradation in ATE RMSE (0.9207m → 1.0061m) and 14.9% degradation in ATE Median (0.7134m → 0.8197m), suggesting challenges in particular scene characteristics. This variation underscores the need for adaptive filtering strategies based on environmental analysis.

### 4.5 Dynamic scene analysis
The quantitative results in Table 1 reveal PCR-ORB's performance characteristics across sequences with varying levels of dynamic content. The YOLOv8 filtering approach demonstrates how different environmental conditions affect trajectory estimation quality.

The YOLOv8 filtering approach demonstrates effectiveness as shown below.

### 4.5.1 Structured environments
Highway driving scenarios show generally positive results in Table 1, where the filtering approach works well with predictable motion patterns and clear object boundaries.

### 4.5.2 Urban scenarios
Moderate urban environments benefit significantly from dynamic object removal, as demonstrated by the substantial improvements in sequences 02 and 04 shown in Table 1.

### 4.5.3 High-dynamic content
Sequences with dense traffic and multiple moving objects show mixed results in Table I, indicating that extremely dynamic scenarios may require additional considerations or adaptive parameter tuning.

### 4.6 Error analysis and system limitations
### 4.6.1 Challenging scenarios
- Dense Traffic: Extremely high dynamic content can challenge the balance between removing dynamic features and preserving sufficient static features for localization.
- Low-Texture Environments: Insufficient environmental texture affects both YOLOv8 detection quality and ORB feature extraction.
- Lighting Variations: While YOLOv8 maintains reasonable detection accuracy across KITTI lighting conditions, extreme shadows or lighting transitions can occasionally impact performance.

### 4.6.2 Filter Robustness
The analysis of Fig. 2 and Table 1 reveals that sequences with moderate levels of dynamic content tend to benefit most from the YOLOv8 filtering approach. The confusion matrices in Fig. 1 show that optimal filtering performance correlates with improved trajectory accuracy in Table 1, particularly evident in sequences 02, 04, and 07.

### 4.7 Computational performance
The YOLOv8 integration maintains real-time performance capabilities, which are essential for practical SLAM applications. The GPU-accelerated implementation provides efficient processing of point cloud filtering while preserving the computational advantages of the original ORB-SLAM3 framework.

### 4.8 Comparison with baseline ORB-SLAM3
The comprehensive results presented in Table 1 demonstrate that YOLOv8-based point cloud filtering provides a viable approach for improving ORB-SLAM3 performance in dynamic environments. While not all sequences show improvements, the significant gains in key scenarios (particularly sequences 02, 04, and 07 as detailed in Table 1) validate the effectiveness of the approach for practical autonomous driving applications.

The mixed results across different sequences highlighted in Table 1 emphasize the importance of environmental context in dynamic SLAM systems and suggest opportunities for future work in adaptive filtering strategies based on scene analysis. The correlation between high detection accuracy shown in Fig. 2 and improved trajectory performance in Table 1 provides evidence for the effectiveness of the YOLOv8-based filtering approach.

## 5 Summary and conclusion
This paper presented PCR-ORB, an enhanced ORB-SLAM3 framework that integrates deep learning-based point cloud refinement for improved operation in dynamic environments. The system addresses the fundamental challenge of dynamic object interference in visual SLAM through a comprehensive multi-stage filtering approach that combines semantic understanding with geometric and temporal constraints.

### 5.1 Key technical achievements
The primary technical achievements of PCR-ORB encompass several significant advances in dynamic SLAM capabilities. The seamless integration of YOLOv8-based semantic segmentation into the ORB-SLAM3 framework demonstrates that modern deep learning techniques can be effectively incorporated into established SLAM systems without compromising core functionality or real-time performance requirements.

The multi-stage filtering pipeline represents a comprehensive approach to dynamic object handling that goes beyond simple semantic classification. By combining semantic information with geometric constraints, temporal consistency analysis, and motion pattern recognition, the system achieves robust dynamic object removal while preserving essential static features necessary for accurate localization.

The CUDA-accelerated implementation proves that sophisticated deep learning inference can be integrated into real-time SLAM systems without exceeding practical computational constraints. The parallel processing architecture effectively distributes computational load across available GPU resources while maintaining synchronization with the main SLAM processing pipeline.

### 5.2 Experimental insights and performance analysis
The experimental evaluation on KITTI dataset sequences 00-09 reveals important insights into the performance characteristics of dynamic object filtering in visual SLAM systems. The results demonstrate that filtering effectiveness varies significantly across different environmental conditions and scene types, indicating that dynamic SLAM approaches must be designed with adaptability and robustness as primary considerations.

Sequence 04 achieved the most significant improvements with 25.9% improvement in ATE RMSE and 30.4% improvement in ATE median, demonstrating that the filtering approach is particularly effective under certain environmental conditions. The characteristics of this sequence, including lighting conditions, scene structure, and dynamic object distribution, appear to be well-suited to the proposed filtering strategy.

The mixed results across different sequences highlight the complexity of dynamic environments and the challenges inherent in developing universally effective filtering approaches. Sequences 05 and 06 showed performance degradation in certain metrics, indicating that over-filtering or incorrect classification of static elements as dynamic can negatively impact localization accuracy.

### 5.3 Limitations and future work

While PCR-ORB demonstrates meaningful improvements in specific scenarios, several limitations warrant acknowledgment and future investigation. The dependency on GPU acceleration for optimal performance limits deployment to platforms with sufficient computational resources, potentially excluding some embedded applications with strict power or cost constraints.

The reliance on YOLOv8 for semantic segmentation introduces dependency on the model's training data coverage and generalization capabilities. Environments or object types not well-represented in training data may experience reduced detection accuracy, affecting overall filtering effectiveness.

Future research directions include extending the system to handle indoor environments and different sensor configurations. Investigation of more efficient neural network architectures could reduce computational requirements while maintaining accuracy. Integration of additional semantic information beyond dynamic object detection could further improve performance.

### 5.4 Broader impact and applications

The development of PCR-ORB has significant implications for autonomous robotics applications requiring robust localization in dynamic environments. The improvements demonstrated in specific scenarios make the system suitable for deployment in autonomous vehicles, mobile robots, and unmanned aerial vehicles operating in complex real-world environments. The real-time performance characteristics and accuracy improvements support practical deployment in safety-critical applications.

The comprehensive evaluation framework and implementation considerations facilitate adoption by the robotics research community. The detailed performance analysis provides insights into the challenges and opportunities in dynamic SLAM development, supporting future research directions. The open discussion of limitations and failure cases contributes to honest scientific discourse about the current state of dynamic SLAM capabilities.

The integration approach demonstrated in PCR-ORB provides a template for incorporating advanced AI techniques into established robotics frameworks. The principles of modular design, backward compatibility, and performance optimization can guide similar integration efforts in other robotics applications. The balance between innovation and practical deployment considerations offers valuable lessons for translating research advances into operational systems.

### 5.5 Concluding remarks

PCR-ORB represents a significant step forward in addressing the challenges of dynamic object interference in visual SLAM systems. The integration of deep learning-based semantic understanding with traditional geometric constraints demonstrates a promising approach for improving SLAM robustness in real-world environments. While the results show scenario-dependent effectiveness rather than universal improvements, the insights gained contribute valuable understanding to the field of dynamic SLAM research.

The technical achievements in real-time deep learning integration, multi-stage filtering design, and CUDA acceleration provide a foundation for future developments in AI-enhanced SLAM systems. The comprehensive evaluation methodology and honest assessment of limitations establish a baseline for measuring progress in dynamic SLAM capabilities.

The work demonstrates that meaningful improvements in dynamic SLAM performance are achievable through careful integration of modern AI techniques with established robotics frameworks. The balance between innovation and practical considerations provides a model for advancing robotics capabilities while maintaining deployment viability. The identification of future research directions offers guidance for continued progress in this important area of robotics research.

As autonomous systems become increasingly prevalent in society, the development of robust dynamic SLAM capabilities becomes essential for ensuring safe and effective operation. The work presented in PCR-ORB provides both technical contributions and research insights that support continued progress toward this important goal, ultimately benefiting both the robotics research community and society.

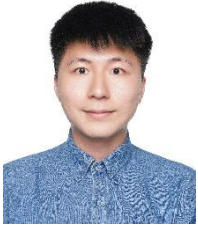

**SHENG-KAI CHEN** received the B.S. degree from Yuan Ze University, Taoyuan, Taiwan in 2024 and pursuing the M.S. degree in 2025 estimated.
He is currently working on the master's degree at Yuan Ze University. In the past, he served as the intern consultant in the digital transformation team of KPMG Taiwan and has done some projects of XAI (e.g. using SHAP, LIT and so on) and solving digital transformation problems of the corporations. Now he is doing research on SLAM improvement, robotic arm model training and compressing the models.
His research interests include visual SLAM, explainable AI and deep learning.

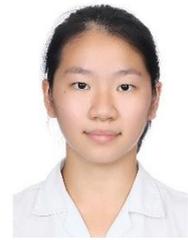

**JIE-YU CHAO** is currently pursuing a B.S. degree in Electrical Engineering at Yuan Ze University, Taoyuan, Taiwan, and is expected to graduate in 2027. She is currently conducting her undergraduate thesis on digital twin technology, with the aim of bridging the gap between virtual and physical systems to reduce deployment costs and enhance system optimization. She has hands-on experience with CUDA and Python, which she uses to accelerate AI-related computations and develop simulation environments for robotics. Her research interests include digital twins, virtual-real system integration, robotics, and cost-efficient digital transformation.

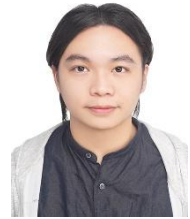

**JR-YU CHANG** is currently pursuing the B.S. degree at Yuan Ze University, Taoyuan, Taiwan, where he enrolled in 2023 and is expected to graduate in 2027. He is now working on his undergraduate thesis, focusing on digital twins. His research aims to reduce the gap between the digital and real world, thereby lowering the cost of system deployment and optimization. He has experience using ROS (Robot Operating System), CUDA, and Python for developing robotic applications, simulation environments, and accelerating computations in AI-related tasks.
His research interests include digital twins, virtual-real system integration, robotics, and cost-efficient digital transformation.

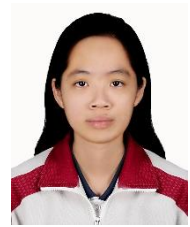

**PO-LIEN WU** is currently pursuing a B.S. degree in Electrical Engineering at Yuan Ze University, Taoyuan, Taiwan, and is expected to graduate in 2027. Her current research focuses on model compression and digital twin technology. Her research interests include efficient model compression and optimization, real-time predictive systems, and the development of digital twin applications.

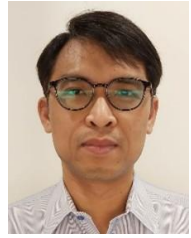

**PO-CHIANG LIN** received the B.S. degree in communication engineering from National Chiao Tung University, Hsinchu, Taiwan, R.O.C. in 1996, and the M.S. and Ph.D. degrees in communication engineering from National Taiwan University, Taipei, Taiwan, R.O.C. in 2005 and 2010, respectively. From 1998 to 2002, he was a system engineer with KG Telecom and SYSTEX Corporation. He is an assistant professor in the Department of Electrical Engineering, Yuan Ze University, Taoyuan, Taiwan, R.O.C.. His research interests include Artificial Intelligence, Robotics, Computer Vision, Wireless Mbile Communication Networks, and STEM Education. From 2018 to 2022, he served as the deputy director of the Science Education Research Center of Yuan Ze University. He served as Wireless Communications and Networking (WCN) Technical Committee (TC) Vice-Chair of the Asia-Pacific Signal and Information Processing Association (APSIPA) from December 2019 to December 2021, and served as WCN TC Chair of APSIPA from January 2022 to December 2023.